\documentclass[10pt,twocolumn,letterpaper]{article}

\usepackage[pagenumbers]{wacv} 

\usepackage{graphicx}
\usepackage{amsmath}
\usepackage{amssymb}
\usepackage{booktabs}
\usepackage{graphicx}
\usepackage{booktabs}
\usepackage{colortbl}
\usepackage{wrapfig}
\usepackage{amsfonts}
\usepackage{algorithm}
\usepackage{enumitem}
\usepackage{algorithmic}

\usepackage[pagebackref,breaklinks,colorlinks]{hyperref}
\usepackage{orcidlink}
\usepackage{hyperref}
\usepackage{soul}

\usepackage[capitalize]{cleveref}
\crefname{section}{Sec.}{Secs.}
\Crefname{section}{Section}{Sections}
\Crefname{table}{Table}{Tables}
\crefname{table}{Tab.}{Tabs.}

\def\wacvPaperID{1289} 
\def\confName{WACV}
\def\confYear{2025}

\usepackage{amsmath,amsfonts,bm}
\usepackage{tikz}

\newcommand*\circled[1]{\tikz[baseline=(char.base)]{
            \node[shape=circle,draw,inner sep=0.2pt, fill=black, font=\footnotesize, text height=1.5ex, text depth=0.1ex, text=white] (char) {#1};}}
            
\def\figref#1{figure~\ref{#1}}

\def\eqref#1{equation~\ref{#1}}

\def\1{\bm{1}}

\def\vc{{\bm{c}}}

\def\vw{{\bm{w}}}
\def\vx{{\bm{x}}}
\def\vy{{\bm{y}}}

\DeclareMathAlphabet{\mathsfit}{\encodingdefault}{\sfdefault}{m}{sl}
\SetMathAlphabet{\mathsfit}{bold}{\encodingdefault}{\sfdefault}{bx}{n}

\DeclareMathOperator*{\argmax}{arg\,max}

\begin{document}
\title{Platypose: Calibrated Zero-Shot Multi-Hypothesis 3D Human Motion Estimation} 

\newcommand{\inst}[1]{$^{#1}$}

\author{
Pawe{\l} A. Pierzchlewicz\inst{1,2},\ 
Caio O. da Silva\inst{2},\ 
R. James Cotton\inst{3,4},\ 
Fabian H. Sinz\inst{1,2,5,6}
\\~\\
$^1$Institute for Bioinformatics and Medical Informatics, Tübingen University, Tübingen, Germany\\
$^2$Department of Computer Science, Göttingen University, Göttingen, Germany\\
$^3$Shirley Ryan AbilityLab, Chicago, IL, USA
Department of Physical Medicine and Rehabilitation,\\ $^4$Northwestern University, Evanston, IL, USA \\
$^5$Department of Neuroscience, Baylor College of Medicine, Houston, TX, USA\\
$^6$Center for Neuroscience and Artificial Intelligence, Baylor College of Medicine, Houston, TX, USA\\
{\tt\small \{ppierzc, sinz\}@cs.uni-goettingen.de}
}



\maketitle

\begin{abstract}
    Single camera 3D pose estimation is an ill-defined problem due to inherent ambiguities from depth, occlusion or keypoint noise. 
    Multi-hypothesis pose estimation accounts for this uncertainty by providing multiple 3D poses consistent with the 2D measurements.
    Current research has predominantly concentrated on generating multiple hypotheses for single frame static pose estimation or single hypothesis motion estimation.
    In this study we focus on the new task of \textit{multi-hypothesis motion estimation}.
    Multi-hypothesis motion estimation is not simply multi-hypothesis pose estimation applied to multiple frames, which would ignore temporal correlation across frames. 
    Instead, it requires distributions which are capable of generating temporally consistent samples, which is significantly more challenging than multi-hypothesis pose estimation or single-hypothesis motion estimation.
    To this end, we introduce Platypose, a framework that uses a diffusion model pretrained on 3D human motion sequences for zero-shot 3D pose sequence estimation.
    Platypose outperforms baseline methods on multiple hypotheses for motion estimation.
    Additionally, Platypose also achieves state-of-the-art calibration and competitive joint error when tested on static poses from Human3.6M, MPI-INF-3DHP and 3DPW.
    Finally, because it is zero-shot, our method generalizes flexibly to different settings such as multi-camera inference
    \footnote{The code is available at \url{https://github.com/sinzlab/platypose}}.
\end{abstract}

\section{Introduction}
\label{sec:intro}
Estimating 3D human motions holds paramount significance across various domains such as gait analysis \cite{Wang2024}, sports analytics \cite{Jiang2022golfpose, Bridgeman2019}, and character animation \cite{Kumarapu2021AnimePose, wang2022markerless}.
In these contexts, motion plays a pivotal role as these applications rely heavily on temporal dynamics.

\textbf{Motion estimation} involves predicting a consistent sequence of poses based on 2D observations, as opposed to \textbf{pose estimation} which estimates a static pose for a single frame.
Despite some methodologies delving into motion estimation \cite{zheng20213d, motionbert2022, motionagformer2024}, a significant challenge persists: current approaches typically provide only a single plausible sequence, thereby neglecting the inherent ambiguity in motion estimation.
This ambiguity stems from various factors, including depth perception limitations, occlusions, and noise in 2D keypoint detection.
Single-hypothesis motion estimation generally outperforms pose estimation by leveraging temporal information.
However, the transition to multi-hypothesis motion estimation, which accounts for multiple possible interpretations of movement, introduces substantial challenges and significantly increases the complexity of the task.
Despite its potential to address the ambiguity problem, multi-hypothesis motion estimation remains largely unexplored. Consequently, the fundamental issue of ambiguity in motion estimation continues to be an unresolved challenge in the field.

\begin{figure}[t]
  \centering
   \includegraphics[width=\linewidth]{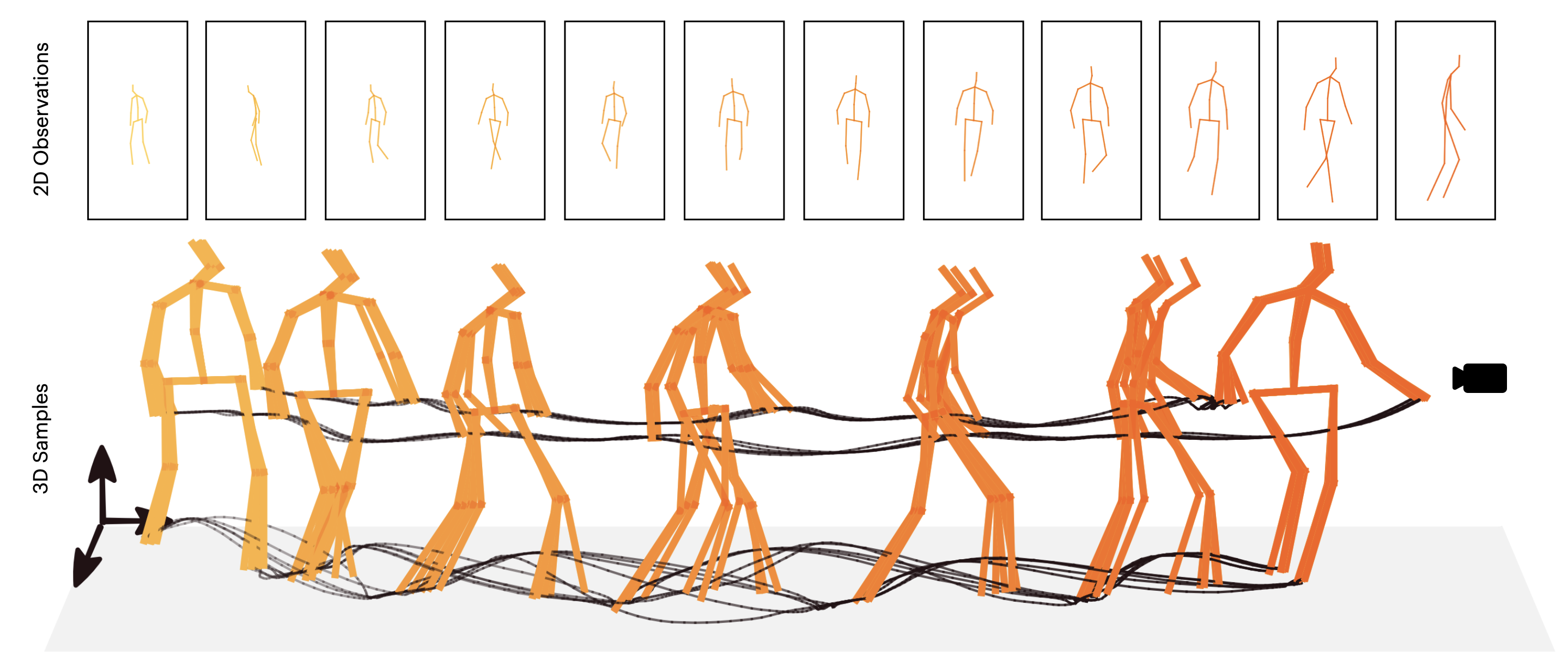}

   \caption{Example samples from the posterior, Platypose generates samples with smooth motion. Darker color indicates later frames in time. Trajectories of wrists and feet are shown for each frame. Top 5 samples are shown at frames 64, 96, 128, 192, 224, 255. Camera icon indicates the direction from which the 2D observations are obtained, thus the depth axis is shown, where increased variance is expected.}
   \label{fig:teaser}
\end{figure}

Incorporating uncertainties into estimates offers valuable insights for users:
medical practitioners, for instance, can benefit from a more transparent and dependable system that alerts them to areas of uncertainty.
Character animators gain a wider range of plausible 3D motions, enhancing their ability to express their artistic vision.
These benefits are not available with the single-hypothesis solutions.
However, estimating multiple hypotheses for \textit{motion} presents a significant challenge compared to multi-hypothesis \textit{pose} estimation (\figref{fig:toy}).
Unlike the latter, multi-hypothesis motion estimation requires temporal consistency, thus increasing the complexity of sampling plausible sequences beyond simply sampling static poses for each individual frame.
Furthermore, many existing multi-hypothesis pose estimation methods suffer from miscalibration, rendering their uncertainty estimates uninformative as they fail to capture the underlying ambiguities of the problem \cite{Pierzchlewicz2022-tu}.

Platypose addresses these challenges as a zero-shot multi-hypothesis motion estimation framework. It infers 3D motion sequences from 2D observations (\figref{fig:teaser}) without requiring explicit training on 2D-3D pairs. This zero-shot capability stems from pretraining a motion diffusion model \cite{Tevet2022-iy} and employing energy guidance \cite{Pierzchlewicz2023-lt}. Consequently, Platypose zero-shot generalizes to new datasets and seamlessly integrates additional data, such as multiple camera inputs, without the need for training separate models for specific camera configurations.

Our key contributions are the following:
\begin{itemize}
    \item We propose a zero-shot 3D motion estimation framework, which uses a motion diffusion model pretrained only on 3D motions to synthesize 3D motions from 2D observations using energy guidance.
    \item We achieve a 10x reduction in inference time through the generation of samples in just 8 steps.
    \item We demonstrate state-of-the-art performance in multi-hypothesis motion estimation, alongside achieving state-of-the-art calibration for pose estimation.
\end{itemize}

\section{Problem Setting}
\label{sec:problem}
\begin{figure}[t]
  \centering
   \includegraphics[width=\linewidth]{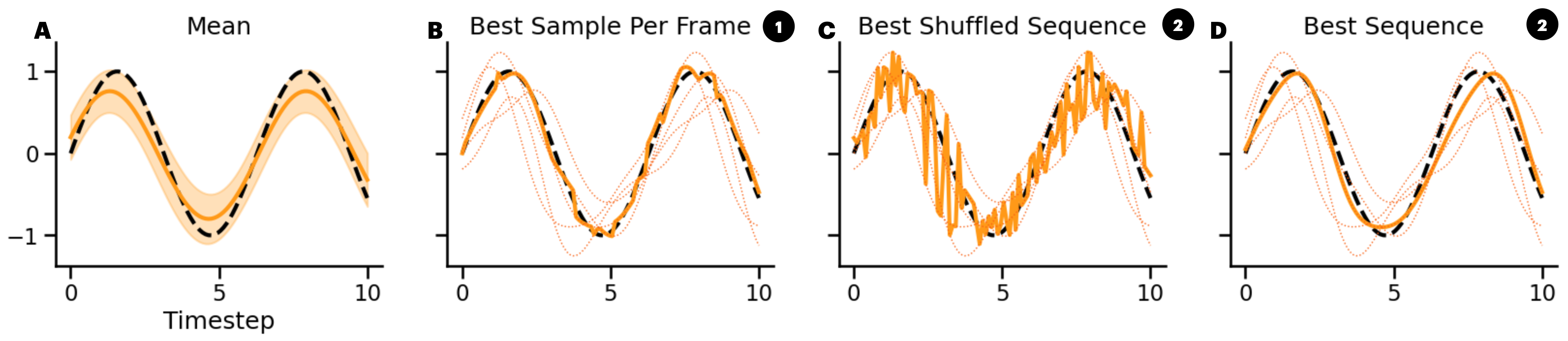}

   \caption{Simplified sequence estimation problem. \textbf{A}) Mean and standard deviation of a Gaussian process fit to a sine function. \textbf{B}) Result of strategy 1 -- choosing the best sample in each frame -- same for both shuffled and non shuffled sequences. \textbf{C}) Result for strategy 2 -- best sequence fit as a whole -- for the shuffled sequences. \textbf{D}) Result for strategy 2 -- best sequence fit as a whole -- for the sequences sampled from the Gaussian process. Dotted lines are the samples from the Gaussian process, solid line is the selected sequence. Dashed line is the ground truth sine wave.}
   \label{fig:toy}
\end{figure}

Estimating multiple hypotheses for motion sequences presents a novel challenge for human behaviour analysis.
The primary goal is to infer the posterior distribution $p(\vx \mid \vy)$ where $\vx$ represents 3D motion sequences and $\vy$ denotes 2D observations of these motions.
Multi-hypotheses motion estimation comes with two central challenges:
\circled{i} Motion estimation entails a significantly higher dimensionality compared to pose estimation.
In pose estimation, $\vx \in \mathbb{R}^{J\times3}$, where $J$ is the number of joints.
For motion estimation $\vx$ expands to $\mathbb{R}^{F\times J\times3}$, where $F$ represents the number of frames.
\circled{ii} Unlike single-hypothesis motion estimation, which predicts a single point estimate such as the mean of the posterior $p(\vx \mid \vy)$, multi-hypothesis motion estimation needs to capture the complex temporal covariance structure.
Consequently, each sample drawn from the distribution should be a valid motion sequence. 
Simply sampling independent poses for each frame overlooks this problem, resulting in an unrealistically noisy motion sequence.
We illustrate this disparity in a simplified scenario (also see~\figref{fig:toy}).

Consider the task of estimating a sine function from noisy observation $f(x) = \sin(x) + \varepsilon$, $\varepsilon \sim \mathcal{N}(0, 0.05)$ (see~\figref{fig:toy}). 
We employ a Gaussian Process with an exponential sine squared kernel fitted to the noisy observations.
We consider sequences sampled from the Gaussian process and sequences where samples are shuffled within each frame, effectively removing temporal correlations.
We consider 2 evaluation strategies.
\circled{1} choosing the best sample independently for each frame and \circled{2} choosing the best sequence as a whole.
Strategy \circled{1} corresponds to pose estimation, while strategy \circled{2} corresponds to motion estimation.
For strategy \circled{1} both non-shuffled and shuffled sequences result in the same outcome, as temporal correlations are not relevant for this strategy.
This observation demonstrates that even when sequence estimates are suboptimal, the per-frame metrics yield low errors.
However, for strategy \circled{2} the best shuffled sequence performs significantly worse than the best non-shuffled sequence.
This simple example demonstrates that even a subpar sequence model can achieve low errors in pose estimation, while motion estimation necessitates a good sequence model capable of sampling consistent -- temporally correlated -- sequences.

\section{Related Work}

\subsection{Multi-hypothesis Learning Based Lifting}
Pose estimation can be approached through either image-to-3D or 2D-to-3D methodologies, the latter is commonly referred to as \textit{lifting}.
\cite{Martinez2017-te} coined the term "lifting" as a \textit{learning-based} task which learns the mapping between 2D and 3D keypoints via a linear ResNet model to transform 2D poses into 3D representations, which outperformed image-to-3D models at that time.
Since then, motion based single-hypothesis methods like PoseFormer \cite{zheng20213d} MotionBERT \cite{motionbert2022} or MotionAGFormer \cite{motionagformer2024} have been dominant. 
They use transformer based architectures to predict single sequences of poses from 2D observations.
However, the 2D-to-3D lifting task presents an inherent challenge as an ill-posed problem.
To address this challenge several multi-hypothesis learning based approaches have been proposed.  \cite{Li2019-wt} and \cite{Oikarinen2021-lg} propose the use of mixture density networks to capture the distribution of plausible poses. 
Meanwhile, \cite{Sharma2019-nh} leverage a variational autoencoder to sample plausible poses and employ ordinal ranking to resolve depth ambiguity. Furthermore, \cite{Kolotouros2021-si}, \cite{Wehrbein2021-gr}, and \cite{Pierzchlewicz2022-tu} use normalizing flows to model the inverse nature of lifting.
Recent advancements employ a diffusion model conditioned on 2D observations \cite{gong2023diffpose, Holmquist2022-ks, Ci2022-sk}. 
The methods above are learning-based methods, thus they require training on 2D-to-3D pairs, which is different from Platypose which is never trained on 2D-to-3D pairs.


\subsection{Zero-Shot Lifting with Diffusion Models}
Diffusion models are a family of generative models designed to invert a diffusion process, by iteratively removing noise \cite{Sohl-Dickstein2015-gy, Ho2020-yx}.
They are capable of sampling from complex distributions, including images \cite{Ho2020-yx, Dhariwal2021-ze}, videos \cite{ho2022video, bartal2024lumiere} or human motions \cite{Tevet2022-iy, raab2024single}, proving highly effective in the realm of human pose estimation \cite{Holmquist2022-ks, gong2023diffpose, Ci2022-sk}.
Moreover, diffusion models have found application in \textit{zero-shot} pose estimation.
ZeDO \cite{Jiang2023-aw} employs a score-based diffusion model with an optimizer in the loop to achieve zero-shot pose estimation.
Initially, a pose is optimized by sampling from the training set and refining its rotation and translation.
Subsequently, this initialized pose undergoes processing via a score-based diffusion model, with an optimizer in the loop to minimize the reprojection error over 1000 steps.
Similarly, PADS~\cite{Ji2024-xi} adopts a comparable strategy to ZeDO, albeit with a distinction in the initialization of the pose.
They further find that the optimization process can be truncated to 450 steps, thus improving sampling speed.
The above methods differ from Platypose, which 1) is designed for \textit{motion} estimation, 2) predicts the denoised sample directly, allowing sample generation in 8 steps, which dramatically decreases inference time and 3) ZeDO and PADs use pose initialization, while Platypose does not.

\subsection{Score Guidance in Human Motion Estimation}
Recent advancements in human motion estimation have leveraged score guidance techniques to enhance the performance of diffusion models. \cite{Zhang2023} employed classifier-free guidance with a graph convolutional network for human mesh recovery, utilizing score guidance to prevent scene penetration. \cite{Rempe2023-cb} applied score guidance to control pedestrian motion alignment with specific objective like waypoint reaching. \cite{Zhang2024-ne} integrated score guidance into their PoseNet diffusion model to refine motion quality, addressing issues such as foot sliding and improving 2D projection adherence.
While these methods primarily use classifier-free guided diffusion at their core and score guidance as a supportive mechanism.
Our approach stands out by employing a purely score-guided method.

\subsection{Human Motion Synthesis}
The synthesis of human motion sequences has gathered significant attention in the recent years.
Now these methods allow high fidelity modeling of the distributions of 3D motions.
Advancements in text-to-motion synthesis have showcased remarkable progress.
\cite{Guo_2022_CVPR} propose a Variational Autoencoder (VAE) mapping text embeddings to a Gaussian distribution in the latent space.
\cite{tevet2022motionclip} expand the text-image embedding space of CLIP \cite{Radford2021LearningTV} to include motion representations.
Building on this, \cite{Tevet2022-iy} have proposed a motion diffusion model (MDM), which is a transformer-based diffusion model that learns the posterior distribution of 3D motions given text descriptions.
Recently, \cite{guo2023momask} propose a motion residual vector quantized VAE achieving the current state-of-the-art performance in motion synthesis.
It is worth noting that the motion synthesis methods model the distribution of motion, but do not solve the motion estimation from 2D observations task.

\subsection{Miscalibration in Human Pose Estimation}
Calibration poses a significant challenge in multi-hypothesis pose estimation, an issue that has been explored in recent studies \cite{Pierzchlewicz2022-tu, Gu2023-gt}.
\cite{Pierzchlewicz2022-tu} highlighted that multi-hypothesis pose estimation methods often suffer from significant miscalibration.
They show that miscalibration tends to erroneously reduce joint errors by underestimating the uncertainty.
Consequently, such miscalibrated methods prove ineffective  to provide insight into the underlying ambiguities in real-world scenarios.



\section{Method}
\subsection{Lifting as Zero-Shot Sampling}
Our goal is to generate samples from a posterior $p(\vx \mid \vy)$ given a prior $p(\vx)$ and a likelihood $p(\vy \mid \vx)$, where $\vx \in \mathbb{R}^{F\times J \times 3}$ are the 3D motion sequences and $\vy \in  \mathbb{R}^{F\times J \times 2}$ are the 2D observations in camera frames, with $F$ representing the number of frames and $J$ the number of joints.
We model the prior $p(\vx)$ as a single-step diffusion model \cite{Ho2020-yx}, which shares similarities with a consistency model \cite{Song2023-tw}.
Initially, a standard diffusion model adds noise to the data via the forward stochastic differential equation
\begin{equation}
    d\vx_t = \mu(\vx_t, t)dt + \sigma(t)d\vw,
\end{equation}
where $t \in [0, T]$ is the diffusion timestep, $\mu$ and $\sigma$ are the drift and diffusion coefficients, and $\vw$ denotes the Wiener process.
Subsequently, to denoise a sample $\vx_t$ at time $t$ the diffusion model follows the probability flow ODE \cite{song2021scorebased}:
\begin{figure*}[t]
  \centering
   \includegraphics[width=\linewidth]{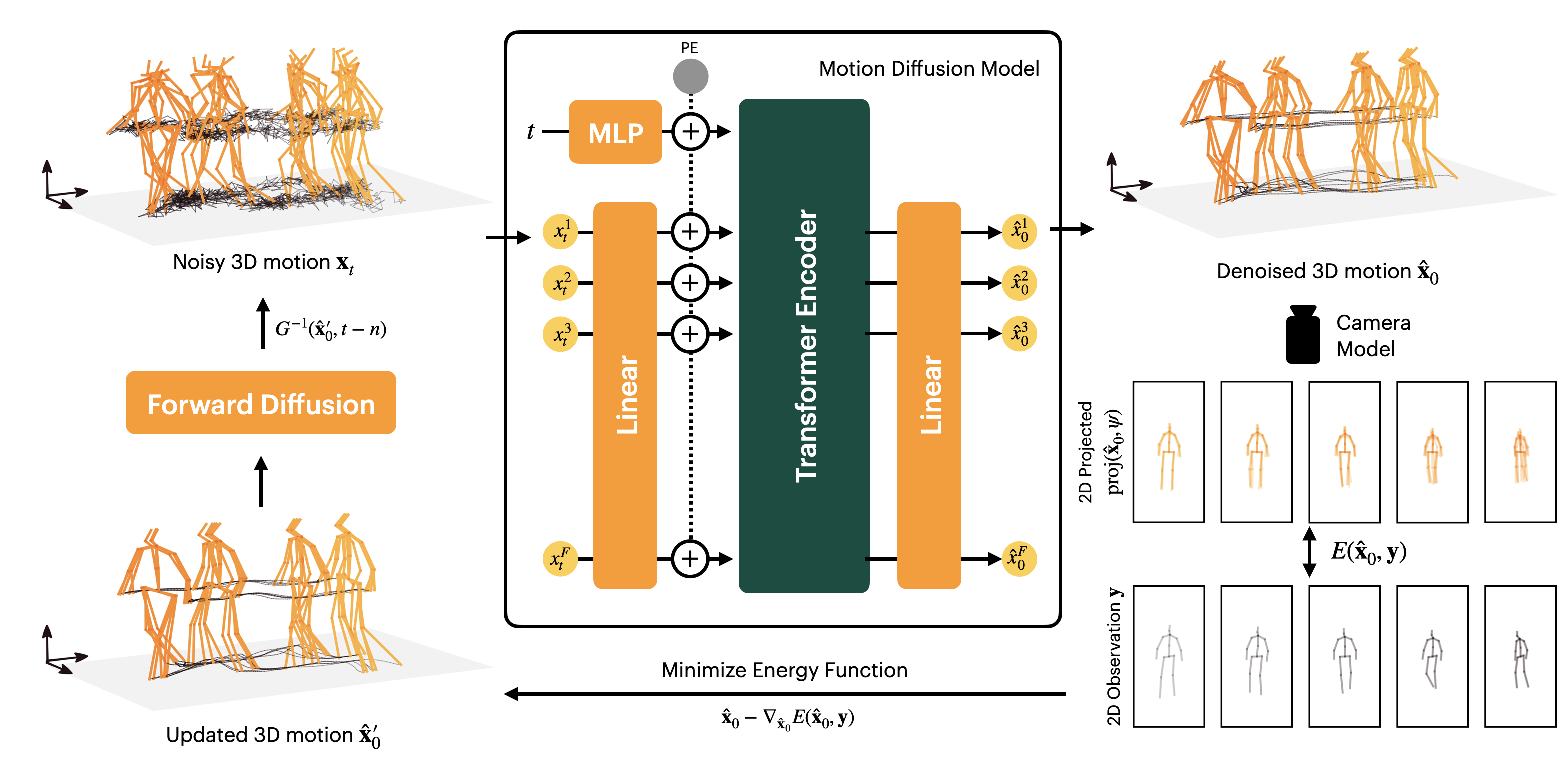}

   \caption{Schematic of sampling using Platypose -- A noisy 3D motion $\vx$ is denoised by a motion diffusion model trained on H36M. The denoised 3D motion samples $\hat{\vx}_0$ are projected to 2D with a camera model. The reprojection error between the projections and 2D observations is minimized. The updated 3D motion is diffused to $t - n$ and passed back into the diffusion model.}
   \label{fig:method}
\end{figure*}

\begin{equation}
    g(\vx_{t}, t) = \frac{d\vx}{dt} = \mu(\vx_t, t) - \frac{1}{2}\sigma(t)^2 \nabla_{\vx_{t}} \log p_t(\vx_{t})
\end{equation}
The value of the fully denoised sample $\vx_0$ can be obtained by integrating $g(\vx_t, t)$ from $t$ to $0$ with the initial state $\vx_t$, which results in the denoiser function $G(\vx_t, t) \rightarrow \vx_0$.
\begin{equation}
    \vx_0 = \vx_t + \int_t^0 g(\vx_\tau, \tau) d\tau = G(\vx_t, t)
\end{equation}

\begin{algorithm}
\caption{Sampling}
\label{alg:sampling}
\begin{algorithmic}
\REQUIRE $\vy$ 2D pose measurement, $\vc$ observation confidence, $T$ total diffusion timesteps, $S$ skip timesteps, $n$ respaced step size, $\lambda$ energy scale, $\theta$ model parameters

\STATE Sample $\vx_{init} \sim \mathcal{N}(\mathbf{0}, \mathbf{I})$
\STATE $\vx_{T-S} \gets G^{-1}(\vx_{init}, T-S)$
\FOR{$t \gets T - S$ \text{ to } $0$}
     \STATE $\hat{\vx}_0 \gets G_\theta(\vx_t, t)$
     \FOR{$k$ iterations}
        \STATE $\hat{\vx}_0 \gets \hat{\vx}_0 - \vc \lambda \nabla_{\hat{\vx}_0} E(\hat{\vx}_0, \vy)$
     \ENDFOR
     \STATE $\vx_{t-n} \gets G^{-1}(\hat{\vx}_0, t-n)$
\ENDFOR
\RETURN $\vx_0$
\end{algorithmic}
\end{algorithm}
To obtain a single-step diffusion model, similar to \cite{Ho2020-yx, Tevet2022-iy, Song2023-tw}, we model the denoiser $G_\theta(\vx, t)$ as a deep neural network model with parameters $\theta$.
Unlike conventional score-based diffusion models, which estimate the score $\nabla_{\vx_t}\log p_t(\vx_t)$ at each timestep, our model directly predicts the denoised sample $\vx_0$ (see model details in \cref{sec:diffusion_model}).
\paragraph{Zero-shot conditioning}
However, zero-shot conditioning and editing is not possible with a single-step denoising process \cite{Song2023-tw}.
Instead, we still need to perform few-step sampling\cite{Song2023-tw}.
To obtain an intermediate step $\vx_{t}$ we run the forward diffusion process $G^{-1}(\vx_0, t) = \sqrt{\bar{\alpha}_{t}}\vx_0 + \sqrt{1 - \bar{\alpha}_{t}}$, where $\bar{\alpha}_t = \prod_{i=0}^t 1 - \sigma(i)$.
This yields a multi-step sampling process with a respacing step size $n$.
\begin{equation}
\label{eq:sampling}
    \vx_{t-n} = G^{-1}(G_\theta(\vx_t, t),\ t-n)
\end{equation}
We find it optimal to use 8 steps in the generation procedure (see \cref{sec:ablation}).
To acquire samples from the posterior distribution we guide the diffusion process using energy guidance \cite{Pierzchlewicz2023-lt} (\cref{fig:method}).
We choose the likelihood to be a Normal distribution around the reprojected 3D keypoints in the image $p(\vy \mid \vx) = \mathcal{N}(\vy; \operatorname{proj}(\vx_0, \vartheta), \lambda^{-1}\mathbf{I})$ for which the energy function corresponds to the reprojection error
\begin{equation}
    \label{eq:energy}
    E(\vx_0, \vy) = \log p(\vy \mid \vx) = \lambda\cdot \|\vy - \operatorname{proj}(\vx_0, \vartheta)\|_2^2.
\end{equation}
Here, $\vartheta$ represents camera parameters, $\vy$ denotes 2D observations and $\lambda$ is the energy scale or precision.
Similar to prior work \cite{Jiang2023-aw, Ji2024-xi} we assume the camera parameters to be known beforehand.
Note that the energy is defined for $\vx_0$ but not for intermediate $\vx_t$ at different steps in the diffusion sampling process.
However, since we use a one-step diffusion $\hat \vx_0 = G_\theta(\vx_t, t)$, we approximate $\vx_0$ for the energy function with $G_\theta(\vx_t, t)$.
We integrate energy guidance into \eqref{eq:sampling}.
\begin{align}
    \hat{\vx}_0 &= G_\theta(\vx_t, t)\\
    \label{eq:update}
    \hat{\vx}_0^\prime &= \hat{\vx}_0 - \lambda \nabla_{\hat{\vx}_0} E(\hat{\vx}_0, \vy)\\
    \vx_{t-n} &= G^{-1}(\hat{\vx}_0^\prime,\ t-n)
\end{align}
We find empirically that performing the update step (\eqref{eq:update}) multiple times improves performance.
Performing $k$ update steps can be interpreted as evaluating the dynamics of the sample using a $k$-th order Yoshida integrator \cite{YOSHIDA1990262}.
We also find that skipping the first $S$ diffusion timesteps improves calibration.
The full sampling procedure is defined in \cref{alg:sampling}.



\subsection{Motion Diffusion Prior}
\label{sec:diffusion_model}
We base our motion diffusion prior on the unconstrained motion diffusion model \cite{Tevet2022-iy}, illustrated in \cref{fig:method}.
The model $G_\theta$ is implemented as an encoder-only transformer model \cite{Vaswani2017-hb}.
The diffusion timestep $t$ is first positionally encoded and then projected into 512 dimensions with 2 linear layers with a SiLU activation function \cite{elfwing2017sigmoidweighted} constructing an input token.
Each frame $f$ of the noisy motion sequence $\vx_t^f \in \mathbb{R}^{J\times3}$ is linearly projected into 512 dimensions and added to standard positional embedding \cite{Vaswani2017-hb}.
Each frame serves as a separate input token for the transformer.
Finally, all output tokens, except the first, are linearly decoded into the pose dimension.

\paragraph{Training} The diffusion model is trained with the objective to predict the denoised sequence $\vx_0$ directly.
\begin{equation}
    \mathcal{L}(\theta) = \mathbb{E}_{\vx_0 \sim \mathcal{D}, t \sim [1, T]} \|G_\theta(\vx_t, t) - \vx_0\|_2^2
    \label{eq:loss}
\end{equation}
The diffusion model is trained with $T=50$ timesteps.
In each training iteration we sample a sequence length from a uniform distribution $\mathcal{U}(1, F)$, where $F$ is the max sequence length.
The training process is described by \cref{alg:training}.

\begin{algorithm}
\caption{Training}
\label{alg:training}
\begin{algorithmic}
\REQUIRE $\mathcal{D}$ dataset of 3D motions, initial model parameters $\theta$, learning rate $\eta$
\REPEAT
    \STATE Sample $\vx_0 \sim \mathcal{D}$, and $t \sim \mathcal{U}(1, T)$, and $f \sim \mathcal{U}(1, F)$.
    \STATE $\vx_t \gets G^{-1}(\vx_0^{0:f}, t)$
    \STATE $\mathcal{L}(\theta) \gets \|G_\theta(\vx_t, t) - \vx_0^{0:f}\|_2^2$
    \STATE $\theta \gets \theta - \eta \nabla_\theta \mathcal{L}(\theta)$
\UNTIL{convergence}
\end{algorithmic}
\end{algorithm}

\subsection{2D Observation Confidences}
Platypose can integrate 2D observation confidences $\vc$ into its sampling process.
This is achieved by scaling the gradient of the energy by $\vc$.
Equation \ref{eq:energy} assumes an isotropic Gaussian likelihood for the reprojected keypoints.
Scaling the energy by $\vc$ is equivalent to changing the precision of this likelihood
$
    E(\vx_0, \vy) \sim \log \mathcal{N}(\vy; \operatorname{proj}(\vx_0, \vartheta), \vc^{-1}\lambda^{-1}\mathbf{I})
$
. To estimate post-hoc confidences, we propose a proxy using ground truth 2D observations $\vy^*$.
Here, we define the confidences as $\vc = |\vy^* - \operatorname{proj}(\vx_0, \vartheta)|$.
When using ground truth keypoints we set $\vc = \mathbf{I}$.
We explore the performance implications of including confidences in \cref{sec:ablation}.

\subsection{Energy Scale Decay}
The energy scale $\lambda$ controls the default variance of the likelihood.
However, relying solely on a singular value of $\lambda$ proves inadequate for optimal performance across all scenarios because the variance in 2D observations exhibits a dependency on depth: poses situated farther from the camera yield less variance in 2D compared to those in closer proximity.
To address this inherent variability systematically, we introduce an energy scale decay mechanism.
This decay process involves a reduction in the energy scale by a factor of $0.1$ whenever the energy $E(\vx_0, \vy)$ increases between consecutive update steps (\eqref{eq:update}).

\section{Experiments}
In this section we introduce experimental results for Platypose on Human3.6M, MPI-INF-3DHP and 3DPW.
We first show the results of Platypose on motion estimation in comparison to a baseline method.
Since baselines in this domain do not exist we construct a baseline by adding a Gaussian distribution to the mean prediction of MotionBERT \cite{motionbert2022} and compare to ZeDO \cite{Jiang2023-aw}, a multi-hypothesis pose estimation method which we extended to multi-hypothesis motion estimation.
Platypose can also predict single frames, therefore, we show a comparison to other single-frame methods.
Details about the training and inference can be found in \cref{sec:training_details}.

\subsection{Datasets and Metrics}
\paragraph{Human3.6M} \cite{h36m_pami, IonescuSminchisescu11} (H36M)
The Human3.6M (H36M) dataset comprises 3.6 million frames from four cameras and corresponding 3D poses obtained via high-speed motion capture. It features 11 actors (6 males, 5 females) across 17 scenarios. For training, we use subjects S1, 5, 6, 7, and 8, with evaluation conducted on subjects S9 and S11.

\paragraph{MPI-INF-3DHP} \cite{mono-3dhp2017} (3DHP) is a single-person 3D pose dataset with 1.3 million frames captured in indoor, green screen and outdoor settings, involving 8 actors (4 males, 4 females). The dataset includes diverse actions ranging from simple to dynamic movements such as exercises. Evaluation is performed on the 6 test sequences defined in the dataset, using the 17 H36M keypoints.

\paragraph{3D Poses in the Wild} \cite{vonMarcard2018} (3DPW)
focuses on in-the-wild human poses captured with moving cameras, comprising 60 videos of 18 actors.
We evaluate on the test set of the 3DPW dataset and use the 17 H36M keypoints.

\paragraph{Evaluation Metrics} We report values in a number of metrics. Firstly, minimum mean per joint position error (minMPJPE),  measures the mean Euclidean distance between each joint of a pose, with the best hypothesis value reported, which is a single frame metric. Secondly, a multi-frame metric -- minimum mean per sequence position error (minMPSPE), which measures the mean MPJPE across the sequence of poses and the minimum is selected across the entire sequence instead of individual frames.
Next, we calculate the mean per joint velocity error (MPJVE), representing the L2 error of joint velocities, based on the sequence that minimizes minMPJPE.
We also report the Procrustes-aligned mean per joint position error (PA-MPJPE), which applies rigid alignment post-processing to the predicted poses before computing minMPJPE and the multi-frame counterpart -- Procrustes-aligned mean per sequence position error (PA-MPSPE).
Finally, we assess the expected calibration error (ECE) as defined in \cite{Pierzchlewicz2022-tu}.
ECE is expressed as $\operatorname{ECE} = |q - \omega(q)|$, where $q \in [0, 1]$ denotes quantiles and $\omega(q)$ represents the frequencies with which ground truth 3D keypoints fall into the predicted distribution's given quantile.

\subsection{Results}
\paragraph{Motion Estimation Baseline}
Given the absence of prior work on estimating multiple hypotheses for motion sequences, we establish our own baseline.
We use the off-the-shelf MotionBERT model \cite{motionbert2022} trained on the Human3.6M dataset, which is a learning based model for single-hypothesis motion estimation achieving 37.5 mm MPJPE error with predicted 2D keypoints.
Inspired by \cite{Pierzchlewicz2022-tu} we develop a multi-hypothesis motion estimation variant of the MotionBERT.
We define the posterior as a Gaussian distribution
$
    p(\vx \mid \vy) = \mathcal{N}(\vx; \mu(\vy), \bm{\sigma}^2\mathbf{I})
$
where $\mu(\vy)$ is the output of the MotionBERT model.
The variance $\bm{\sigma}^2$ is learned to maximize the log likelihood of the posterior $\argmax_{\bm{\sigma}} p(\vx^* \mid \vy)$.
The baseline corresponds to the shuffled sequences introduced in \cref{sec:problem}.
We experimented with a temporally correlated baseline, however, we could not achieve good results consistently.

\begin{figure}[t]
  \centering
   \includegraphics[width=1\linewidth]{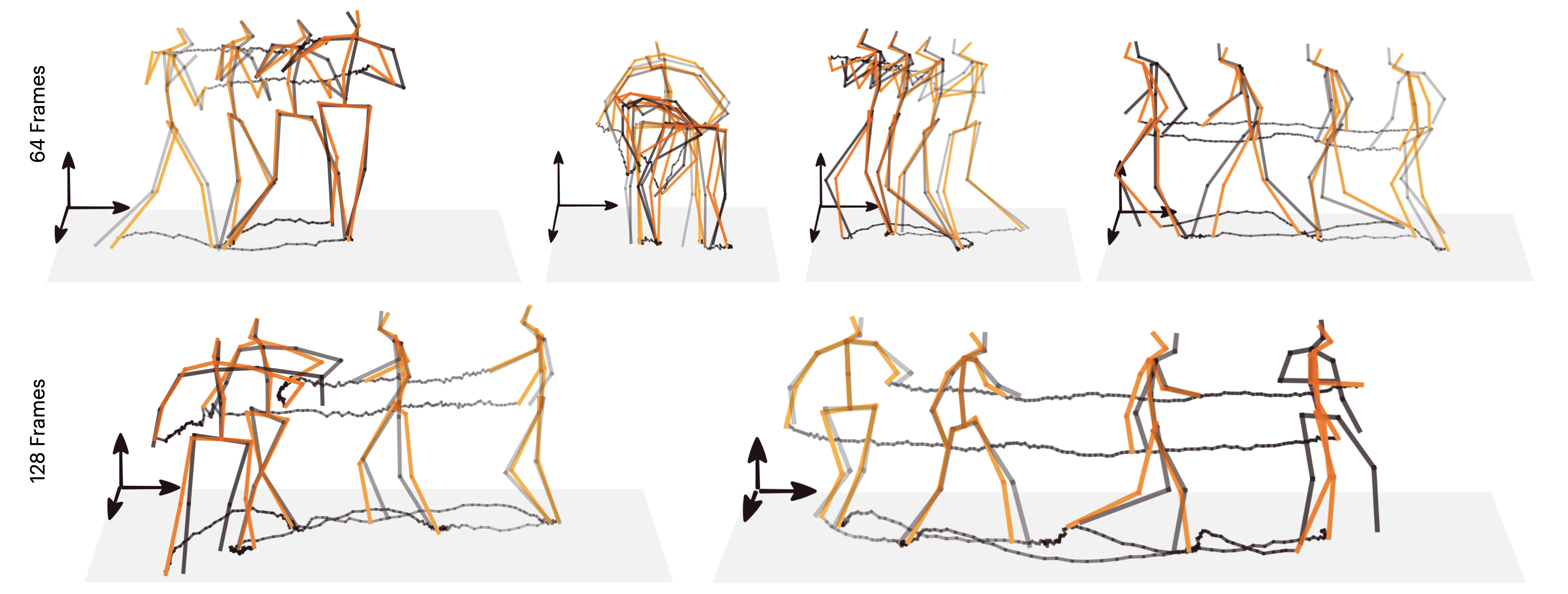}

   \caption{Examples of 3D motion estimates for Human3.6M. Darker color indicates later frames in time. Trajectories of wrists and feet are shown for each frame. Orange poses represent the best sampled hypothesis out of 200 samples, black poses are the ground truth 3D poses. 
   }
   \label{fig:samples}
\end{figure}

\begin{table}[h]
\tiny
\centering
\begin{tabular}{l|c|c|c|c|c}
\toprule
Method & Frames & minMPSPE $\downarrow$ & PA-MPSPE $\downarrow$ & MPJVE $\downarrow$ & ECE $\downarrow$  \\ \midrule
 MotionBERT \cite{motionbert2022}   & 16 & 57.8 & 51.0 & 54.8 &  \textbf{0.03}    \\
              & 64 & 56.4 & 50.4 & 56.0 &  \textbf{0.03}    \\ 
              & 128 & \textbf{56.3} & 50.4 & 56.4 & \textbf{0.04}    \\ \midrule

 Platypose    & 16 & \textbf{47.8} & \textbf{38.5} & \textbf{9.58} &  0.09    \\
              & 64 & \textbf{53.9} & \textbf{43.1} & \textbf{9.67} &  0.09    \\
              & 128 & 60.0 & \textbf{47.7} & \textbf{9.72} &  0.09    \\
\bottomrule
\end{tabular}
\caption{Human3.6M motion estimation results from the baseline MotionBERT + Gaussian noise model and our method, CPN Keypoints, 200 samples, \textbf{bold} indicates best for each number of frames.}
\label{tab:motion_h36m}
\end{table}

\begin{table}[h]
\centering
\tiny
\begin{tabular}{l|c|c|c|c|c}
\toprule
Method & Frames & minMPJPE $\downarrow$ & PA-MPJPE $\downarrow$ & MPJVE $\downarrow$ & ECE $\downarrow$  \\ \midrule
ZeDO$^{\dagger}$ \cite{Jiang2023-aw} & 16  & 31.5 & \textbf{21.7} & \textbf{20.7} &  0.25    \\ 
frame-by-frame    & 64    & \textbf{32.0} & \textbf{22.1} & \textbf{20.4} &  0.24   \\
    & 128              & \textbf{33.7} & \textbf{23.7} & \textbf{22.8} & 0.23    \\
\midrule

 Platypose$^{\dagger}$  & 16   & \textbf{30.6} & 25.6 & 27.4 &  \textbf{0.06}    \\ 
frame-by-frame  & 64 & 32.3 & 26.7 & 30.4 & \textbf{0.05} \\
  & 128 & 34.9 & 28.7 & 33.7 & \textbf{0.03} \\
\midrule 
Method & Frames & minMPSPE $\downarrow$ & PA-MPSPE $\downarrow$ & MPJVE $\downarrow$ & ECE $\downarrow$  \\ \midrule
ZeDO \cite{Jiang2023-aw}& 16   & 178.7 & 103.1 & 252.2 &  0.23    \\ 
evaluated per sequence  & 64 & 224.0   & 125.6 & 309.6 &  0.22    \\ 
& 128  & 240.1 & 133.8 & 328.5 &  0.23    \\
\midrule
  Platypose   & 16   & \textbf{42.5} & \textbf{35.9} & \textbf{2.39} &  \textbf{0.03}    \\ 
evaluated per sequence                & 64   & \textbf{51.5} & \textbf{43.0} & \textbf{2.30} &  \textbf{0.04}    \\ 
                 & 128  & \textbf{59.1} & \textbf{48.5} & \textbf{2.38} &  \textbf{0.07}    \\
\bottomrule
\end{tabular}
\caption{Human3.6M Results with GT Keypoints, 50 samples.
Comparison to ZeDO a single-frame pose estimation method.
${\dagger}$ are methods which generate each frame independently, and the best hypotheses for each frame is selected. The remaining generate the whole sequence and are evaluated by selecting the best sequence as a whole, \textbf{bold} indicates best for each number of frames.}
\label{tab:motion_h36m_gt}
\end{table}

\paragraph{Motion estimation on H36M}
We evaluate the generation of multiple hypotheses for sequences of different lengths (16, 64 and 128 frames) using the H36M dataset (\cref{tab:motion_h36m} -- CPN keypoints; \cref{tab:motion_h36m_gt} -- GT keypoints).
We use $T = 10$, $S = 2$, $\lambda = 30$.
We show examples of the best hypotheses from Platypose in \figref{fig:samples}.
Additional examples can be found in the supplementary materials.
Our baseline model, MotionBERT, reveals that merely adding Gaussian noise to a solid mean estimate is not adequate for achieving high-quality, temporally-consistent, multi-hypothesis sequence estimates.
We demonstrate that Platypose surpasses MotionBERT in terms of minMPJPE and PA-MPJPE, while also significantly outperforming MotionBERT in MPJVE.
The baseline shows lower ECE, which is expected as the variance was explicitly trained for uncertainty quantification.
In \cref{tab:motion_h36m_gt} we compare Platypose to ZeDO \cite{Jiang2023-aw}.
Firstly, we compare the \textit{frame-by-frame} generation case -- each frame is generated independently and the best hypothesis for each frame is selected -- and the \textit{sequence} generation case -- where the whole sequence is generated.
In this case the single-frame ZeDO performs well on the single frame statistics, however, as expected, performs very poorly when evaluated as a sequence.
This shows that Platypose is capable of generating consistent sequences and the use of a motion prior is necessary.

\begin{table}[t]
\centering
\tiny
\begin{tabular}{c|c|c|c|c|c}
\toprule
Cameras     & Frames & minMPSPE $\downarrow$ & PA-MPSPE $\downarrow$ & MPJVE $\downarrow$ & ECE $\downarrow$  \\ \midrule
2 & 16 & 13.6  & 11.4 & 2.17 & 0.12 \\
3 &  & 7.5 & 6.4 & 1.19 & 0.20 \\
4 &  & 4.7 & 4.1 & 0.79 & 0.29 \\
\midrule
2 & 64 & 14.5 & 12.3 & 2.57 & 0.17 \\
3 &  & 8.0 & 6.9 & 1.38 & 0.24 \\
4 &  & 4.8 & 4.3 & 0.75 & 0.34 \\
\midrule
2 & 128 & 16.5 & 14.2 & 2.72 & 0.20 \\
3 &  & 9.3 & 8.2 & 1.36 & 0.28 \\
4 &  & 4.8 & 4.4 & 0.69 & 0.35 \\
\bottomrule
\end{tabular}
\caption{Multi-camera motion estimation results, results for Platypose using 2-4 cameras and for 16, 64 or 128 frames, used GT 2D Keypoints from the Human3.6M dataset, generated 200 samples for each sequence.}
\label{tab:multi_camera}
\end{table}

\begin{table}[t]
\tiny
\centering
\begin{tabular}{l|c|c|c|c|c}
\toprule
Methods               & ZS & $N$ & minMPJPE $\downarrow$  & PA-MPJPE $\downarrow$ & ECE $\downarrow$  \\ \midrule
Sharma et al. (2019) \cite{Sharma2019-nh}&     & 200 & 46.7      & 37.3 &  0.36    \\
Oikarinen et al. (2020) \cite{Oikarinen2020-kq} &     & 200 & 46.2      & 36.3 &  0.16    \\
Wehrbein et al. (2021) \cite{Wehrbein2021-gr} &     & 200 & 44.3      & 32.4 &  0.18    \\
Pierzchlewicz et al. (2022) \cite{Pierzchlewicz2022-tu}  &   & 200  & 53.0      & 40.7 &  \textbf{0.08}    \\
Holmquist et al. (2022) \cite{Holmquist2022-ks} &     & 200 & \underline{42.9}      & \underline{32.4} &  0.27    \\
GFPose (2023) \cite{Ci2022-sk} &     & 200 & \textbf{35.6}      & \textbf{30.5} &  \underline{0.10}    \\ \midrule
ZeDO (2023) \cite{Jiang2023-aw} & $\checkmark$    & 50  & 51.4      & 42.1 &  0.25    \\ 
PADS (2024) \cite{Ji2024-xi}  & $\checkmark$    & 1 & 54.8      &  44.9 &  n/a    \\ 
\rowcolor{gray!25} \textbf{Platypose} (8 steps)   & $\checkmark$   & 50  & 51.8      & 41.5 &  \underline{0.03}    \\ 
\rowcolor{gray!25} \textbf{Platypose} (8 steps)   & $\checkmark$   & 200  & 45.6      & 36.9 &  \textbf{0.02}    \\ 
\rowcolor{gray!25} \textbf{Platypose} (16 steps)   & $\checkmark$   & 50  & \underline{50.9}      & \underline{40.69} &  0.04    \\ 
\rowcolor{gray!25} \textbf{Platypose} (16 steps)   & $\checkmark$   & 200  & \textbf{45.0}      & \textbf{36.3} &  \underline{0.03}    \\
\bottomrule
\end{tabular}
\caption{Human3.6M pose estimation Results, CPN Keypoints, \textbf{bold} is best, \underline{underline} is second best. The best values are considered separately for zero-shot and learning based methods. ZS are zero-shot methods. $N$ -- number of hypotheses.}
\label{tab:cpn_h36m}
\end{table}

\paragraph{Multi-Camera Motion Estimation}
Multi-camera setups can vastly improve the accuracy of motion estimation \cite{Kolotouros2021-si, Li2019-wt}.
Platypose can naturally scale to multiple cameras without any additional training.
By simply modifying the energy function, the model can effectively handle data from multiple viewpoints.
The energy function with observations from $N$ cameras is defined as
$
    E(\vx, \vy) = \sum_i^N \| \vy_i - \operatorname{proj}(\vx, \vartheta_i) \|_2^2
$
. The results presented in \cref{tab:multi_camera} showcase Platypose's performance across varying numbers of cameras (2-4) on the H36M dataset.
Joint errors decrease as the number of cameras increases, yet the distribution tends to become miscalibrated. 
This phenomenon is likely attributed to the increasing rigidity imposed by the constraints from multiple cameras, leading to overconfident estimation.

\paragraph{Pose Estimation on H36M}
We evaluate Platypose on the multi-hypothesis pose estimation task. 
To achieve pose estimation, instead of inputing a sequence of tokens, a single token for the pose is passed into the model.
We use $\lambda = 30$ and $T = 12$, $S = 4$ (\textit{8 Steps}) or $T = 20$, $S = 4$ (\textit{16 steps}).
We consider the standard predicted 2D keypoints from the off-the-shelf cascading pyramid network (CPN) model (CPN, \cref{tab:cpn_h36m}).
Platypose achieves comparable results to ZeDO on 50 samples and significantly outperforms ZeDO on 200 samples. Additionally, Platypose exhibits superior calibration compared to ZeDO.
Furthermore, Platypose surpasses other zero-shot methods and narrows the performance gap between learned methods on predicted keypoints.
Moreover, Platypose demonstrates state-of-the-art calibration.
Thus, showing that even though Platypose was designed to estimate motion it is also capable of doing pose estimation.

\begin{table}[t]
\tiny
\centering
\begin{tabular}{l|c|c|c}
\toprule
Methods               & $N$ & MPJPE $\downarrow$ & ECE $\downarrow$  \\ \midrule
Kanazawa et al. \cite{kanazawa2018endtoend}  & 1 & 113.2 & n/a \\ 
Gong et al. \cite{Gong2021-rw} &   1 & 73.0 & n/a \\ 
Gholami et al. \cite{Gholami_2022_CVPR}  & 1 & 68.3 & n/a \\
Chai et al. \cite{chai2023global} &   1 & \textbf{61.3} & n/a \\ \midrule
Muller et al.$^\dagger$ \cite{Mueller:CVPR:2021} & 1 & 101.2 & n/a \\
ZeDO$^\dagger$ \cite{Jiang2023-aw}   & 50  & 69.9   &  0.28    \\ 
\rowcolor{gray!25} \textbf{Platypose}$^\dagger$ (8 St.)    & 50  & 74.6  &  \underline{0.08}    \\ 
\rowcolor{gray!25} \textbf{Platypose}$^\dagger$ (8 St.)     & 200  & \underline{64.2}    &  \textbf{0.07}    \\
\rowcolor{gray!25} \textbf{Platypose}$^\dagger$ (16 St.)    & 50  & 74.7  &  \underline{0.08}    \\ 
\rowcolor{gray!25} \textbf{Platypose}$^\dagger$ (16 St.)     & 200  & 64.4    &  \underline{0.08}    \\ \bottomrule
\end{tabular}
\caption{3DHP Results, GT Keypoints, \textbf{bold} is best, \underline{underline} is second best, $\dagger$ are zero-shot methods. $N$ -- number of hypotheses. \textit{St.} stands for Steps.}
\label{tab:3DHP}
\end{table}

\paragraph{Cross-Dataset Pose Estimation}
In this section, we assess Platypose's ability to generalize across datasets, as shown in \cref{tab:3DHP,tab:3DPW}.
Using our pretrained diffusion prior from the H36M dataset, we evaluate Platypose's performance on both the 3DHP and 3DPW test sets.
We use $\lambda = 10$ and $T = 12$, $S = 4$ (\textit{8 Steps}) or $T = 20$, $S = 4$ (\textit{16 steps}).
Our analysis reveals that Platypose exhibits better calibration compared to alternative methods.
Platypose performs well on the 3DPW dataset, where it outperforms previous zero-shot and learned methods in both minMPJPE and PA-MPJPE metrics.
This highlights Platypose's robustness and adaptability across diverse datasets, indicating its potential for real-world applications.

\begin{table}[h]
\tiny
\centering
\begin{tabular}{l|c|c|c|c}
\toprule
Methods               & $N$ & MPJPE $\downarrow$ & PMPJPE $\downarrow$ & ECE $\downarrow$  \\ \midrule
Kocabas et al. \cite{kocabas2019vibe} & 1 & 93.5 &  56.5 & n/a \\ 
Kocabas et al. \cite{Kocabas_PARE_2021} & 1 & 82.0 &  50.9 & n/a \\ 
Gong et al. \cite{Gong2021-rw}  & 1 & 94.1 & 58.5 & n/a \\ 
Gholami et al. \cite{Gholami_2022_CVPR} & 1 & 81.2 & 46.5 & n/a \\
Chai et al. \cite{chai2023global} & 1 & 87.7 & 55.3 & n/a \\ \midrule
ZeDO$^\dagger$  \cite{Jiang2023-aw}   & 1  & 69.7   &  40.3 & n/a    \\ 
\rowcolor{gray!25} \textbf{Platypose}$^\dagger$ (8 St.)   & 50  & 60.1    & 39.6 & \underline{0.05}    \\ 
\rowcolor{gray!25} \textbf{Platypose}$^\dagger$ (8 St.)  & 200  & \underline{50.6}    & \underline{34.2} & \textbf{0.04}    \\
\rowcolor{gray!25} \textbf{Platypose}$^\dagger$ (16 St.)   & 50  & 60.2    & 38.7 & 0.06    \\ 
\rowcolor{gray!25} \textbf{Platypose}$^\dagger$ (16 St.)  & 200  & \underline{50.8}    & \textbf{33.9} & \underline{0.05}    \\
\bottomrule
\end{tabular}
\caption{3DPW Results, GT Keypoints, \textbf{bold} is best, \underline{underline} is second best, $\dagger$ are zero-shot methods. $N$ -- number of hypotheses. \textit{St.} stands for Steps.}
\label{tab:3DPW}
\end{table}

\paragraph{Inference Speed Comparison}
When testing against ZeDO on a GeForce 2080 Ti we find that Platypose generates a sample in 1.1s, which is 10x faster than ZeDOs 11s. This provides a significant boost in performance, allowing real-time generation of samples using Platypose (\cref{tab:inference_speed}).

\paragraph{Single hypothesis estimation lacks significance for calibrated models}
For $n=1$, the MPJPE on H36M is 141.6 mm, while the geometric median yields 98.1 mm. This aligns with expectations, as calibrated distributions rarely produce low errors for single samples \cite{Pierzchlewicz2022-tu}. Current methods often exhibit overconfidence \cite{Pierzchlewicz2022-tu}, trading calibrated uncertainty for increased precision. Given the limited significance of such results, we have omitted them from our tables.

\subsection{Ablation Study}
\label{sec:ablation}
\begin{figure}[t]
  \centering
   \includegraphics[width=\linewidth]{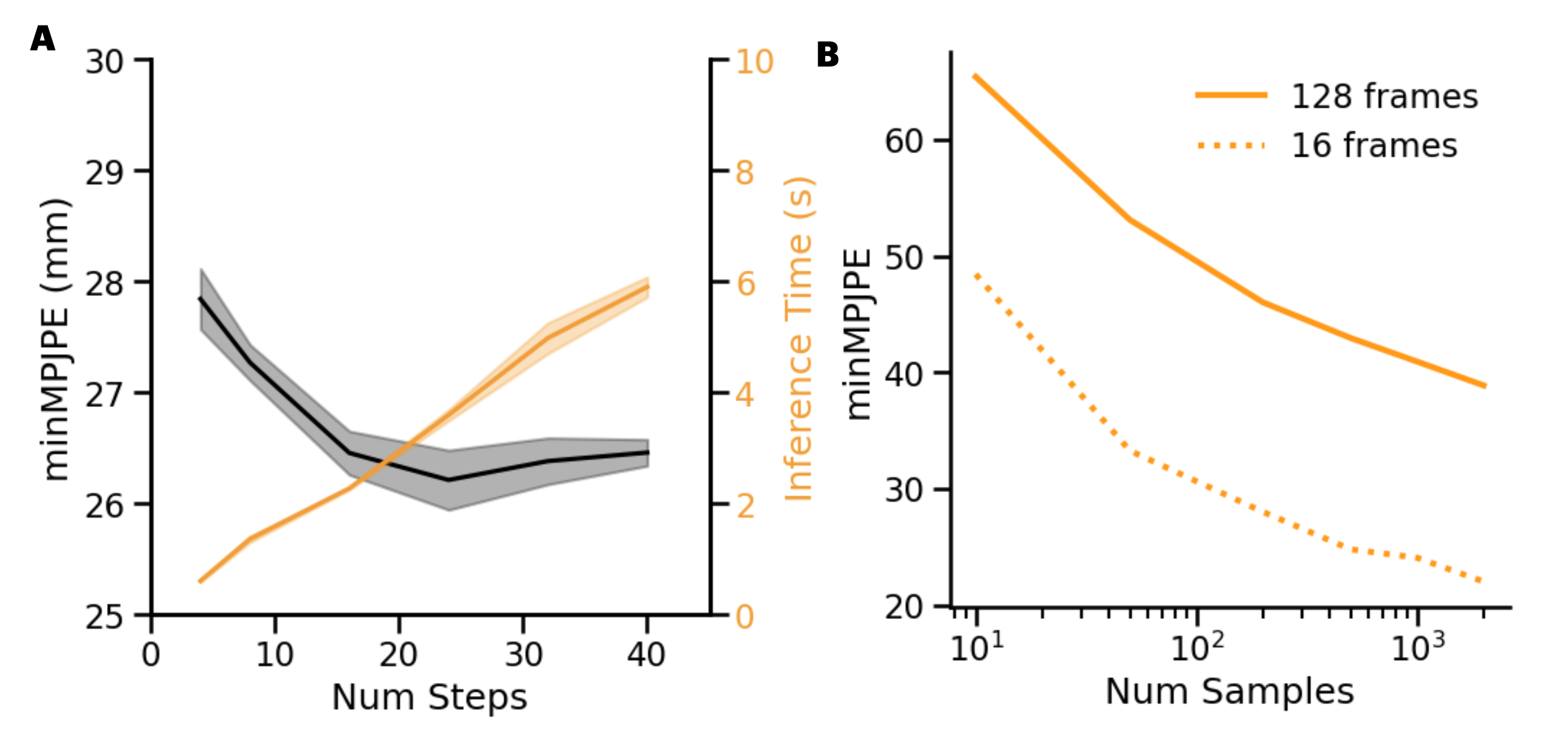}

   \caption{\textbf{A}) Impact of the number of diffusion steps on minMPJPE and the inference time. Evaluated for single frame estimation. Mean and standard deviation are plotted from 3 seeds. \textbf{B}) Impact of the number of samples on minMPJPE for two different sequence lengths.}
   \label{fig:ablation}
\end{figure}
\paragraph{Influence of inference steps}
We evaluate how the number of inference steps affects model performance.
While the model was originally trained on 50 steps, we can change the number of steps and quantify the effect on performance and inference speed. 
In \cref{fig:ablation}a we show the results of different numbers of inference steps.
We evaluate on 1 frame long sequences, initiating sampling from 20\% of the diffusion timesteps. For instance, with 8 inference steps, sampling starts from step 2 and progresses to step 10. We observe a slight performance enhancement with increased steps, albeit at a notable expense in inference time.

\paragraph{Number of hypotheses}
We evaluate how changing the number of sampled hypotheses impacts joint error.
As expected, increasing the number of hypotheses decreases the error.
Longer sequences necessitate more samples for comparable performance to shorter ones.
Sampling long sequences requires sampling from higher dimensional spaces.
This is because more samples are needed to cover the same volume as for the short sequences.
We show the results in \cref{fig:ablation}b.
Evaluation is conducted on every 10th example of the H36M test set using GT keypoints.

\paragraph{Influence of confidence}
Including the confidence of 2D keypoints should intuitively improve performance (\cref{tab:confidence}).
In this ablation study we compare how 2D observation confidences affect performance.
We test on single frames, with and without confidence estimates on H36M.
We find that including confidence estimates helps decrease errors, but has no significant impact on calibration.
Further improving the method of estimating the 2D confidence could lead to better performance.


\paragraph{Energy scale decay}
We assess the impact of energy scale decay through an ablation study, focusing on its effect on the minMPJPE error in the 3DPW dataset. Our findings reveal that implementing energy scale decay leads to a 0.8 mm reduction in error, as shown in \cref{tab:energy_decay}.

\section{Limitations}
Although Platypose demonstrates strong performance, it is not without limitations.
We outline these limitations below.
\circled{1} Like other zero-shot methods, Platypose relies on accurate camera parameters for estimating 3D poses.
Additionally, it assumes prior knowledge of the root trajectory in 3D space.
\circled{2} Platypose is not optimized for single hypothesis estimation.
While it may not excel in this scenario, it is important to note that it is not its primary function.
In the case of a well calibrated distribution, a single sample is not likely to fall close to the ground truth 3D pose.
Thus, we would argue that a good performance of a multi-hypothesis method on one sample indicates an overconfident distribution.
\circled{3} There are instances where Platypose fails to generate reasonable 3D hypotheses.
These failures may stem from issues with 2D keypoint detection or unexplained ambiguities.
We include videos and figures showcasing these failure examples in the supplementary materials.

\section{Conclusions}
In this paper we introduce Platypose, a zero-shot framework for estimating 3D human motion sequences from 2D observations.
To the best of our knowledge we are the first to tackle the problem of multi-hypothesis motion estimation.
We condition a pretrained motion diffusion model using energy guidance to synthesize plausible 3D human motion sequences given 2D observations.
We achieve state-of-the-art performance in comparison to baseline methods.
Furthermore, Platypose proves to be also capable of well-calibrated pose estimation.

\subsection*{Acknowledgments}
We thank Arne Nix and John Peiffer for their helpful feedback and discussions.
Funded by the German Federal Ministry for Economic Affairs and Climate Action (FKZ ZF4076506AW9) and  SFB 1456 Mathematics of Experiment project number 432680300. RJC is supported by the Research Accelerator Program of the Shirley Ryan AbilityLab and the Restore Center P2C (NIH P2CHD101913).

%
%
\bibliographystyle{splncs04}
\bibliography{main, paperpile}

\clearpage

\renewcommand{\figurename}{Supplementary Figure}
\renewcommand{\tablename}{Supplementary Table}
\renewcommand*{\thetable}{S\arabic{table}}
\renewcommand*{\thefigure}{S\arabic{figure}}
\setcounter{figure}{0}
\setcounter{table}{0}

\appendix

\section{Supplementary Materials}

\subsection{Training and Inference Details}
\label{sec:training_details}
The diffusion prior was trained for 600,000 steps using the AdamW \cite{Loshchilov2017} optimizer with a learning rate of $10^{-4}$ and batch size of 64. 
Training was executed on a single GeForce 2080Ti GPU in 25 hours.
The model was trained only on the train set of Human3.6M with a max sequence length of $F=256$ and $T=50$ diffusion timesteps.
We train using the H36M only to maintain a fair comparison to previous methods.

\subsection{Pose Initialization}
As shown in \cite{Jiang2023-aw, Ji2024-xi}, pose estimation using diffusion models can benefit from first initializing the pose. 
\cite{Ji2024-xi} initializes the pose by inverse projection
\begin{equation}
    \vx_{init} = \frac{K^{-1}\vy}{\|K^{-1}\vy\|_2} \|T\|_2,
\end{equation}
where $K$ is the camera matrix, $T$ is the root trajectory and $\vy$ is the 2D observation.
We compare the methods using the proposed initialization.

\subsubsection{Influence of initialization}
We compare 3 different initialization strategies.
Firstly, using a random initialization from a standard normal, secondly using the inverse projection and finally using the ground truth 3D pose (oracle).
We find (\cref{tab:init}) that using the inverse projection initialization for single frames improves the performance mariginally and impairs calibration.
Furthermore, by using the ground truth we could further improve the performance ($1.5$mm$\downarrow$) with a smaller decrease in calibration.
This indicates that improving the initialization strategy can leave room for further improvements using Platypose, however, given the tradeoff with calibration we choose not to use initialization.

\begin{table}[h]
\centering
\begin{tabular}{l|c|c}
\toprule
Initialization    & minMPJPE & ECE\\
\midrule
Gaussian      & 39.5  & 0.04       \\
Inv. Proj.      & 39.3  & 0.08       \\
Oracle      & 38.0  & 0.06     \\
\bottomrule
\end{tabular}
\caption{Impact of the initialization strategy on minMPJPE and calibration. Tested on single frames.}
\label{tab:init}
\end{table}

\subsection{Additional Platypose Evaluation on Ground Truth Keypoints}
We additionally evaluate the performance of Platypose on estimating 3D poses from ground truth keypoints (\cref{tab:gt_h36m}) and 3D motions from ground truth keypoints under two evaluation strategies -- \textit{per frame} and \textit{per sequence} (\cref{tab:motion_h36m_gt_200}).
We use the same setup as for the CPN keypoints.
We find that also in this case Platypose is well calibrated and achieves comparable performance to the uncalibrated ZeDO on pose estmation.

\begin{table}[h]
\centering
\scriptsize
\begin{tabular}{l|c|c|c|c|c}
\toprule
Methods               & ZS & $N$ & minMPJPE $\downarrow$  & PA-MPJPE $\downarrow$ & ECE $\downarrow$  \\ \midrule
SemGCN      &  & 1 & 43.8      & - &  n/a    \\
Oikarinen et al. &     & 200 & 31.8      & 26.3 &  0.16    \\
Kolotouros et al.  &   & 200  & 37.1      & - &  0.07    \\ \midrule
ZeDO  & $\checkmark$    & 50  & 37.0      & 27.5 &  0.25    \\ 
PADS   & $\checkmark$    & 1 & 41.5      &  33.1 &  n/a    \\ 
\rowcolor{gray!25} \textbf{Platypose} (8 Steps)   & $\checkmark$   & 50  & 39.1      & 32.8 &  \textbf{0.04}    \\ 
\rowcolor{gray!25} \textbf{Platypose} (8 Steps)  & $\checkmark$   & 200  & \underline{31.9}      & \underline{27.6} &  \textbf{0.04}    \\
\rowcolor{gray!25} \textbf{Platypose} (16 Steps)   & $\checkmark$   & 50  & 37.7      & 31.8 &  \underline{0.05}    \\ 
\rowcolor{gray!25} \textbf{Platypose} (16 Steps)  & $\checkmark$   & 200  & \textbf{30.9}      & \textbf{26.8} &  \underline{0.05}    \\
\bottomrule
\end{tabular}
\caption{Human3.6M Results, GT Keypoints, \textbf{bold} is best, \underline{underline} is second best, ZS are zero-shot methods. $N$ -- number of hypotheses.}
\label{tab:gt_h36m}
\end{table}

\begin{table}[h]
\centering
\tiny
\begin{tabular}{l|c|c|c|c|c|c}
\toprule
Method & Samples & Frames & minMPSPE $\downarrow$ & PA-MPSPE $\downarrow$ & MPJVE $\downarrow$ & ECE $\downarrow$  \\ \midrule
 Platypose  & 50 & 16   & 41.5 & 35.1 & 4.96 &  0.03    \\ 
evaluated per frame                && 64   & 45.0 & 38.6 & 5.20 &  0.04    \\ 
                 && 128  & 47.1 & 40.7 & 5.26 &  0.07    \\
\midrule
Platypose  & 200 & 16   & 34.7 & 29.7 & 5.22 &  0.05    \\ 
evaluated per frame                && 64   & 37.9 & 32.8 & 5.55 &  0.05    \\ 
                 && 128  & 40.1 & 34.7 & 5.43 &  0.09    \\
    \midrule
Platypose & 200 & 16   & 35.8 & 30.7 & 2.29 &  0.05    \\ 
evaluated per sequence   && 64               & 45.2 & 37.6 & 2.21 &  0.05   \\ 
    && 128              & 52.9 & 43.3 & 2.27 &  0.09    \\
\bottomrule
\end{tabular}
\caption{Human3.6M Results with GT Keypoints.
Different Platypose sequence evaluation methods and number of hypotheses.
The whole sequence is sampled and is evaluated by either selecting the best sequence as a whole -- \textit{evaluated per sequence} -- or by selecting the best hypothesis for each frame -- \textit{evaluated per frame}.}
\label{tab:motion_h36m_gt_200}
\end{table}

\subsection{Calibration of the Multi-Camera Setup}
\begin{figure}[t]
  \centering
   \includegraphics[width=\linewidth]{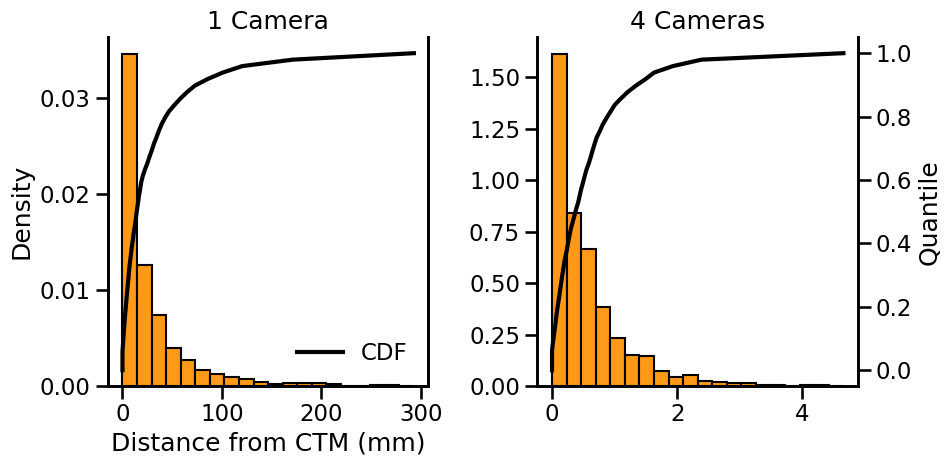}

   \caption{Error histograms from the center tendency measure (CTM) for calibration for estimates from 1 or 4 cameras. The cumulative distribution function for each distribution is plotted.} 
   \label{fig:error_hist}
\end{figure}

We investigate the increasing ECE in the Multi-Camera setup.
To demonstrate where this effect comes from we compare the distances from the central tendency measure of each distribution for 1 and 4 cameras (\cref{fig:error_hist}).
We find that the error distribution is approximately 100x narrower in the case of 4 cameras.
As a result the cumulative distribution function (CDF) becomes very steep for the 4 cameras case.
Thus, small deviations in the mean prediction will result in substantial changes in the quantile assignments.
Thus, it becomes increasing difficult to reliably measure calibration at such precision.

\subsection{Alternative optimization objectives}
Although the L2 objective has an elegant probabilistic interpretation, the performance could be further improved with alternative objectives.
One such objective is the Geman-McClure penalty loss.
The Geman-McClure penalty loss shows minor improvements to the minMPJPE of 0.7mm on the 3DPW dataset \cref{tab:Geman}.

\begin{equation}
    \mathcal{L}_{GM} = \frac{||x - x^*||_2^2}{s^2 + ||x - x^*||_2^2}
\end{equation}

\begin{table}[t!]
    \centering
    \begin{tabular}{c|c}
        \toprule
        \textbf{Method} & \textbf{minMPJPE (mm)}\\
        \midrule
        L2 & 64.2\\
        Geman-McClure & 63.5\\
        \bottomrule
    \end{tabular}
    \caption{Ablation of using Geman Mc-Clure penalty loss. Tested on the 3DPW dataset.}
    \label{tab:Geman}
    
\end{table}

\subsection{Failure case analysis and additional examples}
\begin{figure*}[t]
  \centering
   \includegraphics[width=0.8\linewidth]{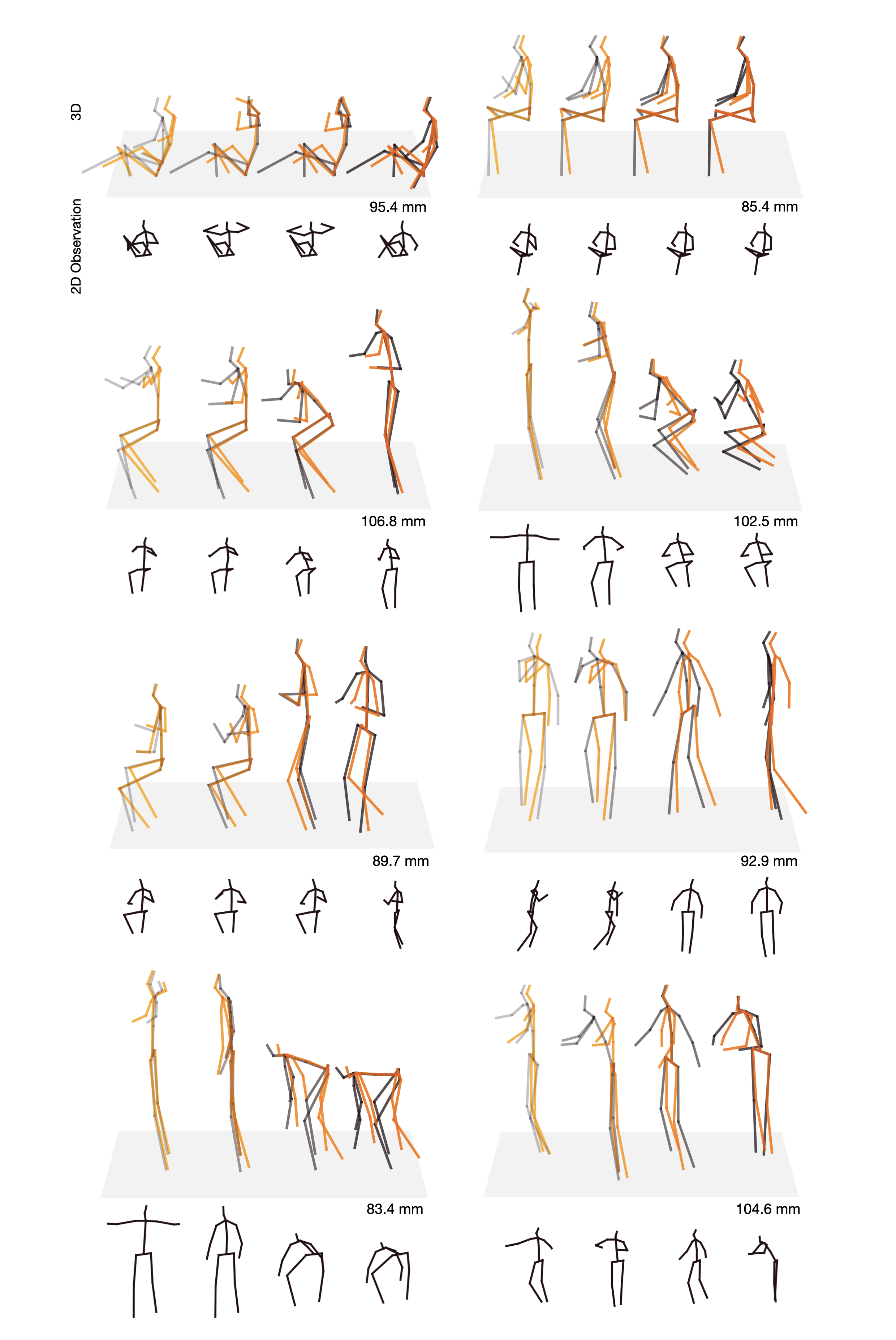}
   \caption{Failure cases -- visualization of failure cases, where MPSPE $>$ 80mm. Orange are samples from Platypose and black are ground truth 2D and 3D poses. We show frames 0, 42, 84 and 127. The MPSPE of these examples are displayed.}
   \label{fig:failure}
\end{figure*}

\begin{figure*}[t]
  \centering
   \includegraphics[width=0.8\linewidth]{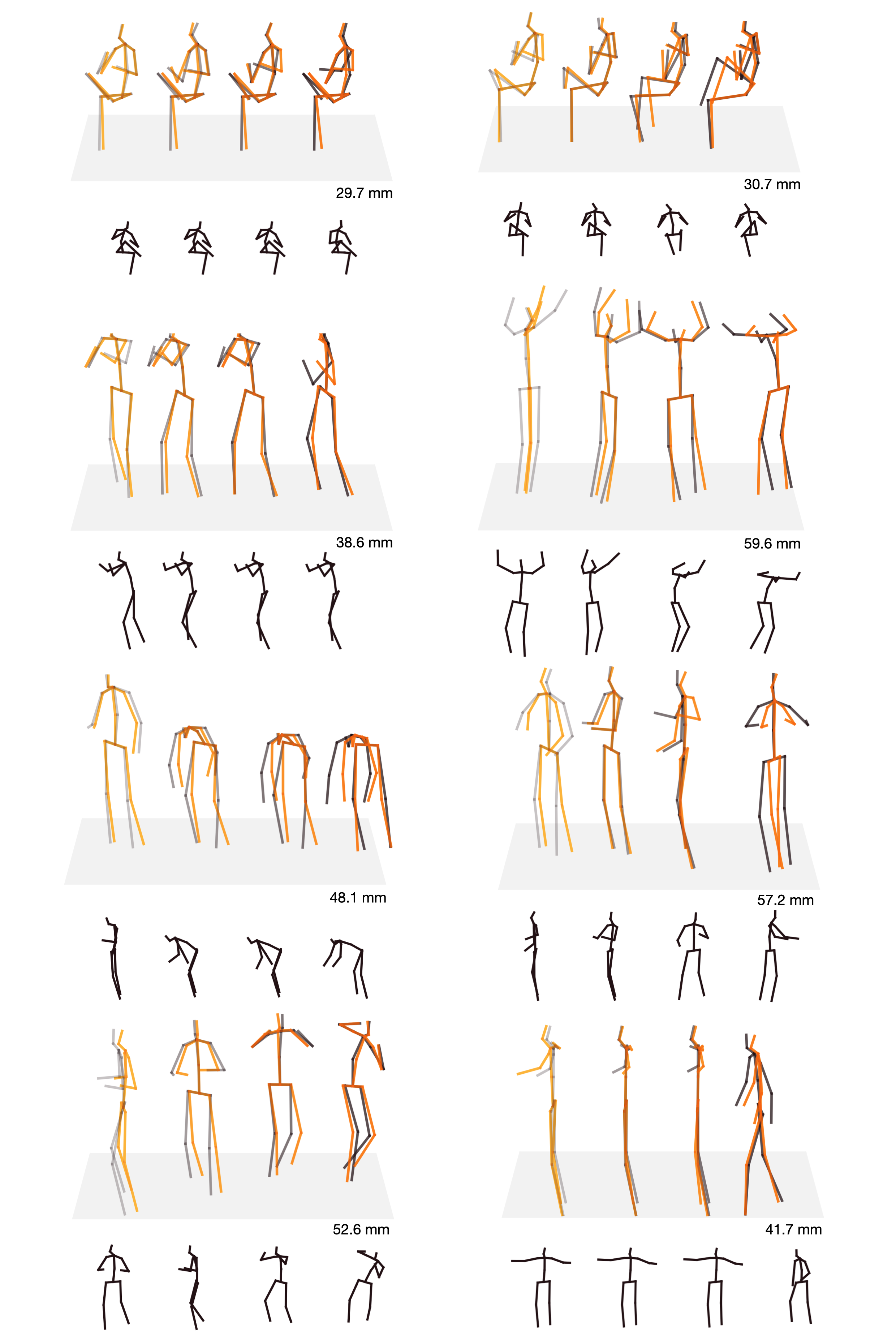}
   \caption{Visualization of samples from Platypose. Orange are samples from Platypose and black are ground truth 2D and 3D poses. We show frames 0, 42, 84 and 127. The MPSPE of these examples are displayed.}
   \label{fig:additional_examples}
\end{figure*}

In \cref{fig:failure} we show some failure cases.
The majority of the failure can be attributed to ambiguous 2D observations where either the pose is not very informative about the 3D pose, e.g. standing sideways to the camera, or the 3D pose is a difficult pose like crouching or sitting.
If there is a mixture of easy and difficult poses, the best hypothesis might favour the sequence which best fits to the easy poses and does not fit too well to the more difficult frames.
In \cref{fig:additional_examples} we show additional examples of samples.

\begin{table}[t]
\scriptsize
\centering
\begin{tabular}{l|c|c}
\toprule
Method    & Sample Time (s) & Iterations\\
\midrule
ZeDO      & 11.0  & 1000       \\
\rowcolor{gray!25}Platypose & \textbf{1.1} & \textbf{8}     \\\bottomrule
\end{tabular}
\caption{Time to generate 1 sample on Nvidia GeForce 2080 Ti. \textbf{Bold} indicates best.}
\label{tab:inference_speed}
\end{table}

\begin{table}[t!]
\centering
\vspace{-\topskip}
\begin{tabular}{c|c}
\toprule
Method    & minMPJPE (mm)\\
\midrule
with energy decay & 64.2\\
        w/o energy decay & 65.0\\
\bottomrule
\end{tabular}
\caption{Impact of energy decay on minMPJPE for 3DPW. Tested on pose estimation.}
    \label{tab:energy_decay}
\end{table}

\begin{table}
\centering
\begin{tabular}{l|c|c}

\toprule
Confidence    & minMPJPE & ECE\\
\midrule
     & 46.4  & \textbf{0.02}       \\
\multicolumn{1}{c|}{$\checkmark$}      & \textbf{45.6}  & \textbf{0.02}       \\
\bottomrule
\end{tabular}
\caption{Impact of confidence on minMPJPE and calibration. Tested on 1 frame. \textbf{Bold} indicates best.}
\label{tab:confidence}
\end{table}

\end{document}